\documentclass[conference]{IEEEtran}
\IEEEoverridecommandlockouts

\usepackage{cite}
\usepackage{amsmath,amssymb,amsfonts}

\usepackage{graphicx}
\usepackage{textcomp}
\usepackage{xcolor}
\def\BibTeX{{\rm B\kern-.05em{\sc i\kern-.025em b}\kern-.08em
    T\kern-.1667em\lower.7ex\hbox{E}\kern-.125emX}}

\usepackage{amsmath}
\usepackage{amssymb}
\usepackage{amsthm}
\usepackage{algorithm}
\usepackage{xspace}
\usepackage{graphicx}
\usepackage{subfigure}
\usepackage{wrapfig}
\usepackage{graphicx}
\usepackage{subfigure}
\usepackage{url}
\usepackage{color}
\usepackage{multirow}
\usepackage{gensymb}
\usepackage{longtable}
\usepackage{tikz}
\usepackage{comment}
\usepackage{dsfont}
\usepackage{framed}
\usepackage{enumitem}
\usepackage{xspace}
\usepackage{bm}
\usepackage{pifont}
\usepackage{algpseudocode}
\usepackage{booktabs}

\usepackage{multibib}
\newcites{citenew}{Reference}
\usepackage{hyperref}
\usepackage{makecell} 
\usepackage{mathtools}

\theoremstyle{plain}

\newtheorem{lemma}{Lemma}

\theoremstyle{definition}

\newcommand{\dagalg}{\texttt{IMAX}\xspace}
\newcommand{\dro}{DRO\xspace}
\newcommand{\ouralg}{\texttt{3D-Learning}\xspace}
\newcommand{\kldro}{\texttt{KL-DRO}\xspace}

\newcommand{\wdro}{\texttt{W-DRO}\xspace}
\newcommand{\ml}{\texttt{DFL}\xspace}

\newcommand{\da}{\texttt{DA}\xspace}
\newcommand{\cost}{\textit{Cost}\xspace}
\newcommand{\utility}{\textit{Utility}\xspace}
\newcommand{\pact}{P_{\text{act}}\xspace}
\newcommand{\pidle}{P_{\text{idle}}\xspace}
\newcommand{\DDPM}{\texttt{DDPM}\xspace}
\newcommand{\zinit}{Z_0\xspace}

\algrenewcommand\algorithmicrequire{\textbf{Input:}}
\algrenewcommand\algorithmicensure{\textbf{Output:}}

\begin{document}

\title{3D-Learning: Diffusion-Augmented Distributionally Robust Decision-Focused Learning
\thanks{* Correspondence to: Jianyi Yang (jyang66@uh.edu).}}

\author{\IEEEauthorblockN{ Jiaqi Wen}
\IEEEauthorblockA{
\textit{University of Houston}
}
\and
\IEEEauthorblockN{ Lei Fan}
\IEEEauthorblockA{
\textit{University of Houston}}
\and
\IEEEauthorblockN{Jianyi Yang*}
\IEEEauthorblockA{
\textit{University of Houston}}

}

\maketitle

\begin{abstract}
Predict-then-Optimize (PTO) pipelines are widely employed in computing and networked systems, where Machine Learning (ML) models are used to predict critical contextual information for downstream decision-making tasks such as cloud LLM serving, data center demand response, and edge workload scheduling. However, these ML predictors are often vulnerable to out-of-distribution (OOD) samples at test time, leading to significant decision performance degradation due to large prediction errors.  To address the generalization challenges under OOD conditions, we present the framework of Distributionally Robust Decision-Focused Learning (DR-DFL), which trains ML models to optimize decision performance under the worst-case distribution.
Instead of relying on classical Distributionally Robust Optimization (DRO) techniques, we propose Diffusion-Augmented Distributionally Robust Decision-Focused Learning (\ouralg), which searches for the worst-case distribution within the parameterized space of a diffusion model. By leveraging the powerful distribution modeling capabilities of diffusion models, \ouralg identifies worst-case distributions that remain consistent with real data, achieving a favorable balance between average and worst-case scenarios. Empirical results on an LLM resource provisioning task demonstrate that \ouralg outperforms existing DRO and Data Augmentation methods in OOD generalization performance.\footnote{Source Code Link: \href{https://github.com/CIGLAB-Houston/3DLearning.git}{https://github.com/CIGLAB-Houston/3DLearning.git}}

\end{abstract}

\begin{IEEEkeywords}
Diffusion Models, Distributionally Robust Learning, Decision-Focused Learning.
\end{IEEEkeywords}
\section{Introduction}
Many context-aware decision-making problems in computing and communication networks can be formulated within the Predict-then-Optimize (PTO) framework, where effective decision-making critically depends on the accurate prediction of system context \cite{DFL_survey_mandi2024decision}. One key example is resource provisioning for cloud (Large Language Model) LLM serving, where the accurate token-level workload prediction is essential for efficiently allocating computing resources (e.g., GPU cores or frequency) to mitigate over-provisioning or service degradation \cite{DL_workload_scheduling_ye2024deep,energy_measurement_AI_caspart2022precise,energy_consumption_LLM_heterogeneous_wilkins2024hybrid}. Another critical application is demand response in AI data centers, where accurate prediction of AI workloads of different types and renewable energy availability is essential for scheduling the AI computation workload to reduce energy costs while maintaining service-level agreements \cite{AI-data-center-grid-colangelo2025turning,data_center_demand_response_zhang2021hpc,pricing_demand_response_liu2014pricing}.

In many PTO applications, decision performance is highly sensitive to specific types of prediction errors. For instance, in cloud resource provisioning, underestimating the workload can lead to sever service degradation, whereas overestimation may simply incur additional cost. As a result, the decision performance can be sub-optimal by training ML models solely to minimize prediction error without considering its impact on downstream decision objectives. Decision-Focused Learning (DFL) \cite{end-to-end-learning_donti2017task,DFL_survey_mandi2024decision} addresses the limitation of traditional prediction-focused learning by training ML models in an end-to-end manner to directly optimize the final decision objective. By aligning the learning process with decision performance, DFL can provide more effective and robust strategies. 

Despite its advantages over prediction-focused learning, DFL still struggles to generalize under Out-of-Distribution (OOD) testing scenarios—a common challenge in dynamic ML-based systems. In cloud workload scheduling, for instance, shifts in user demand, task types, or market dynamics can lead to fluctuating workload patterns that differ significantly from those seen during training. When faced with such distribution shifts, a DFL model trained solely on in-distribution data may make decisions with poor performance. This vulnerability arises because standard DFL optimizes decision performance based solely on the empirical training distribution, without accounting for potential distribution shifts at test time. To address this limitation, we introduce the Distributionally Robust Decision-Focused Learning (DR-DFL) framework, which seeks to optimize decision performance under the worst-case distribution within an ambiguity set based on the training data.  With a well-designed ambiguity set, DR-DFL can enable more resilient decision-making under real-world OOD deployment scenarios \cite{DRO_mohajerin2018data,Wasserstein_DRO_kuhn2019wasserstein,KL-DRO_hu2013kullback}.

A central challenge in DR-DFL lies in the modeling of the ambiguity set, which defines a distribution discrepancy measure to capture meaningful variations around the training distribution. However, Distributionally Robust Optimization (DRO) methods with traditional ambiguity modeling often lead to suboptimal DR-DFL performance. Ambiguity sets based on $\phi$-divergences (e.g., KL divergence) restrict the distributions to be absolutely continuous with respect to the training distribution, thereby excluding test distributions with shifted support. As a result, DR-DFL with $\phi$-divergence-based ambiguity sets may yield non-robust solutions under support shift, even when enhanced with data augmentation techniques.  In contrast, Wasserstein distance-based ambiguity sets allow for support variation but often result in intractable optimization problems, particularly when the decision objective is non-convex. In such cases, solving the DRO problem typically requires relaxations, which may result in overly conservative training, ultimately degrading the average-case performance.
 These limitations highlight the need for a more expressive and computationally tractable ambiguity modeling in DR-DFL.

 This paper focuses on addressing the challenges of ambiguity set design and proposes a novel DR-DFL framework based on diffusion models with the following main contributions:
\begin{itemize}
\item \textbf{Diffusion-based Ambiguity Modeling}. We introduce a new ambiguity modeling based on the score matching loss of diffusion models, offering several advantages. First, it allows the ambiguity set to include distributions with diverse and shifted support. Second, by constraining the score matching loss, we can ensure that candidate distributions remain consistent with the underlying data distribution. Finally, it enables the tractable search for the worst-case distribution within the parameterized space of the diffusion model.
\item \textbf{Diffusion-Augmented Algorithm Design}. 
We propose the \textit{\underline{D}iffusion-Augmented \underline{D}istributionally Robust \underline{D}ecision-Focused Learning} (\ouralg) algorithm, which integrates diffusion-based ambiguity modeling into the DR-DFL pipeline. It addresses the challenges of the constrained, non-convex inner maximization problem by combining the dual learning techniques with diffusion policy optimization methods. Furthermore, given the inner maximization output, we design the min-max solver for DR-DFL based on the diffusion sampling.  
\item \textbf{Performance Evaluation on LLM Resource Provisioning}. 
We formulate the resource provisioning problem for LLM inference as a PTO pipeline and evaluate \ouralg against a range of DRO and data augmentation baselines. Simulation results demonstrate that \ouralg significantly outperforms traditional DRO methods and data augmentation approaches in both average-case and worst-case performance across test datasets exhibiting various distribution shifts. Moreover, under various noisy perturbation scenarios, \ouralg demonstrates exceptional noise robustness and stability.

\end{itemize}

\section{Formulation and Appications}

\subsection{Decision-Focused Learning}
\begin{figure}[t]
        \centering
        \includegraphics[width=\linewidth]{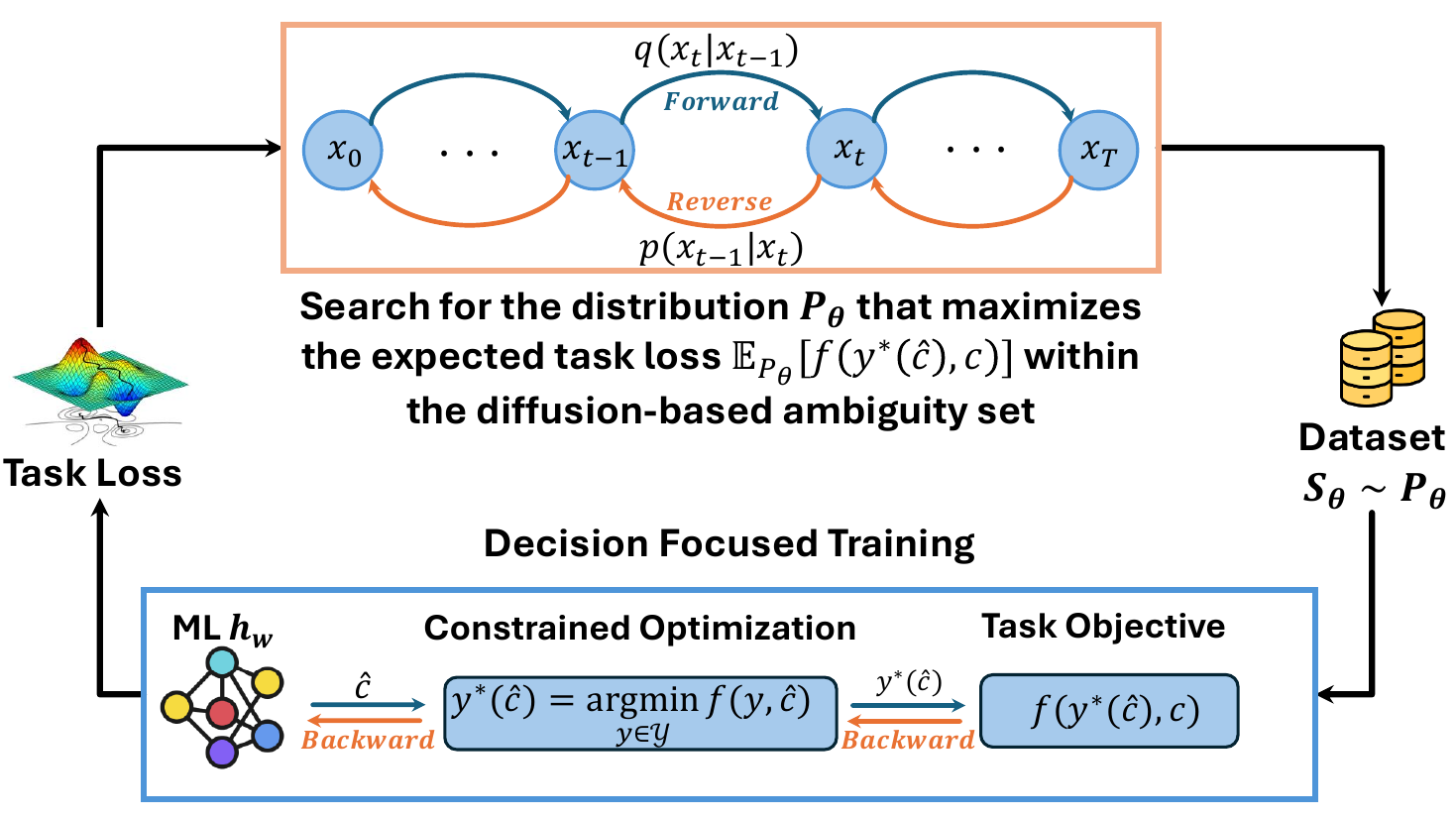} % 改为0.7倍宽
        \caption{Framework of Decision Focused Learning}
        \label{Fig_FW_DFL}    
    \end{figure}

 PTO problems have wide applications in computing and communication networks \cite{PTO_clustering,wang2024end,wu2020proactive}. In a PTO pipeline as shown in \ref{Fig_FW_DFL}, a ML predictor $h_w\in\mathcal{H}$ with weight $w$ maps an input $v$ (e.g. historical context or side information) into a context prediction $\hat{c}$.  Then a decision-making step is taken to optimize the decision objective based on the predicted context $\hat{c}$. We consider a general decision-making objective as
\begin{equation}\label{eqn:optimization}
   y^*(c) =\arg\min_{y\in\mathcal{Y}} f(y,c),
\end{equation}
where $f$ is the objective functions and $\mathcal{Y}$ incorporates the constraints on the decision variable $y$.

In a standard ML training loss, we usually define the training loss by some discrepancy measure $l(\hat{c},c)$ such as Mean Squared Error (MSE) or Cross-Entropy (CE) between the predicted label $\hat{c}$ and the ground-truth label $c$. Denote $x=(v,c)$ as a labeled sample. We minimize the empirical loss $\mathbb{E}_{S_0}[l(h_w(v),c)]$  based on a training dataset $S_0=\{x_1,\cdots, x_{|S_0|}\}$ with $|S_0|$ samples drawn from an underlining distribution $P_0$. 
Existing works \cite{DFL_survey_mandi2024decision} have demonstrated that such standard ML training may not achieve satisfactory decision performance because the information of decision objective \eqref{eqn:optimization} is not incorporated in the ML training.  
Therefore, DFL is proposed to train the ML model in an end-to-end style by directly minimizing the decision objective $f(y^*(\hat{c}),c)$. The training objective of DFL is expressed as
\begin{equation}\label{eqn:DFL}
h_w = \arg\min_{h_w\in \mathcal{H}} \mathbb{E}_{S_0}[f(y^*(\hat{c}),c)],
\end{equation}
where $y^*(\hat{c})$ is the solution of \eqref{eqn:optimization} given a predicted context $\hat{c}=h_w(v)$. DFL is more general than standard ML training because it reduces to a standard ML training by choosing the decision objective as the label discrepancy measures.

\subsection{Distributionally Robust Decision-Focused Learning}
Although DFL directly optimizes the decision objective, it suffers from significant performance drop in the OOD testing environment. As illustrated by examples in Section \ref{sec:applications}, the distribution of testing context can shift a lot over time, resulting in unreliable decision performance. 

DRO is a widely-adopted framework to improve OOD generalization performance.
For the PTO pipeline with the decision objective \eqref{eqn:optimization}, we introduce DR-DFL which trains predictor to minimize the worst-case decision objective by a min-max optimization:
\begin{equation}\label{eqn:DRO}
h_w = \arg\min_{h_w\in \mathcal{H}}\max_{P\in \mathcal{B}(P_0,\epsilon)} \mathbb{E}_{P}[f(y^*(\hat{c}),c)],
\end{equation}
where $P_0$ is the underlining distribution of the training dataset $S_0$, and $\mathcal{B}(P_0,\epsilon)$ is the ambiguity set which contains possible testing distributions and is typically modeled as a distribution ball $\mathcal{B} (P_0,\epsilon)= \left\{ P \mid D(P,P_0) \leq \epsilon \right\}$ given a distribution discrepancy measure $D$ and a budget $\epsilon$. 
Based on the discrepancy measure $D$ in the ambiguity set, we can get different DRO methods. The widely-adopted discrepancy measures include Wasserstein distance and the family of $\phi-$divergence which lead to Wasserstein \dro \cite{DRO_mohajerin2018data,DRL_chen2020distributionally,Wasserstein_DRO_gao2023distributionally,risk_stochastic_opt_zhao2018data} and $\phi-$divergence \dro \cite{KL-DRO_hu2013kullback, DR_BO_divergence_husain2023distributionally, DRBO_kirschner2020distributionally,chance-constrained-jiang2016data} respectively. 

The choice of the ambiguity set has a large effect on the performance of \dro. As the decision objective in the PTO pipeline can be very sensitive to the distribution shifts, it becomes critical to choose a proper ambiguity set in DR-DFL.
 As the focus of this paper, we will discuss the challenges of ambiguity modeling and present our diffusion-based ambiguity modeling in Section \ref{sec:methods}.

\subsection{Applications in Computing and Communication Networks}\label{sec:applications}
The considered PTO problem has wide applications for context-aware decision-making problems in computing and communication networks.
\subsubsection{Workload-Aware Resource Provisioning for LLM Inference} \label{sec:application_LLM_serving}
With the rapid deployment of AI, particularly large language models (LLMs), the substantial energy costs of AI workloads have become a critical concern. In AI data centers, inference workloads often constitute a large fraction of total computing demand \cite{DL_workload_scheduling_ye2024deep,energy_measurement_AI_caspart2022precise,energy_consumption_LLM_heterogeneous_wilkins2024hybrid}. 
As LLMs are increasingly adopted, serving systems must process a large volume of LLM inference requests. Given the limited computing capacity and fluctuating demand, data center operators usually need to provision resources in advance based on predicted workloads \cite{online_resource_provisioning_comden2019online}. However, the distribution of inference demand can evolve significantly, making it difficult to forecast workloads accurately and to strike an effective balance between energy efficiency and performance guarantees based on the prediction.
 We develop a distributionally robust LLM workload predictor focusing on the objectives of serving performance and energy costs with a detailed case study in Section \ref{sec:simulation}.
\subsubsection{Demand Response in AI Data Centers}
The rapid advancement of AI technologies has driven an exponential increase in the demand for high-performance AI data centers, which places a substantial burden on power grids and contributes to high energy costs \cite{report_demand_surge_davenport2024ai}. 
Despite their high power consumption, the flexibility in AI training workloads opens up opportunities for demand response (DR), where data centers adapt their energy usage in response to grid signals such as real-time electricity prices and carbon intensity \cite{AI-data-center-grid-colangelo2025turning,data_center_demand_response_zhang2021hpc,pricing_demand_response_liu2014pricing}. The power usage can be controlled by workload shifting (e.g., deferring non-urgent jobs to off-peak hours) or power capping (e.g., reducing server utilization or GPU frequency). In this way, AI data centers can significantly lower energy costs while satisfying Service Level Agreements (SLAs) for AI workloads.
However, the performance of DR strategies relies on the accurate prediction of the workloads and the grid signals, which is complicated by the high variability of AI workloads and the integration of renewable sources.
Given the context uncertainty, it is essential to develop distributionally robust and decision-focused ML models to provide accurate predictions which directly support DR objectives under uncertainty. 

\subsubsection{Channel-Aware Edge Data Center Selection}
Edge Data Centers (EDCs) offer heterogeneous hardware and software resources—including CPUs, GPUs, memory, and pre-deployed AI models—to support low-latency Mobile Edge Computing (MEC) services. In a typical MEC system, the incoming computational requests must be assigned to suitable EDCs in order to meet service quality objectives including latency and inference accuracy \cite{edge_resource_allocationmeskar2018fair,edge_survey_luo2021resource,AI_edge_computing_tuli2023ai}.
Unlike traditional cloud computing environments, where network latency is relatively stable, edge computing environments are highly dynamic due to user mobility, wireless conditions, and network congestion. This makes it challenging to predict channel conditions which are essential context to optimize the service quality of EDC selection.
Thus, to fight against uncertainty in real-world edge environments, a robust and decision-focused channel quality predictor is critically needed for optimizing the user-to-EDC assignments with their service objectives.

\section{Diffusion Ambiguity Modeling}\label{sec:methods}
\subsection{Challenges in Ambiguity Modeling}

The choice of ambiguity set in DR-DFL \eqref{eqn:DRO} has a large effect on the robustness performance and the solution tractability. The limitations of the commonly-used Wasserstein and $\phi-$divergence based ambiguity sets are introduced as below.

If we choose the discrepancy measure as the Wasserstein distance $D_{\mathrm{W}}(P, P_0)$, we get the Wasserstein \dro (\wdro). 
However, it is difficult to find a tractable solution for \wdro.
Some methods \cite{DRO_mohajerin2018data,DRL_chen2020distributionally,Wasserstein_DRO_gao2023distributionally} reformulate Wasserstein-constrained DRO into a finite optimization based on the assumption of convex objectives which typically do not hold in deep learning. Other methods convert \wdro into an adversarial training with norm constraints on samples \cite{staib2017distributionally,Wasserstein_DRO_gao2023distributionally}, but these solutions can be overly conservative and cannot fully exploit the benefits of probabilistic ambiguity modeling to improve generalization.

Alternatively, the distribution discrepancy can be measured by the family of $\phi$-divergence $D_{\phi}(P\|P_0)=\mathbb{E}_{P_0}[\phi(\frac{\mathrm{d} P}{\mathrm{d} P_0})]$ where $\phi$ is a convex function with $\phi(1)=0$ \cite{KL-DRO_hu2013kullback, DR_BO_divergence_husain2023distributionally, DRBO_kirschner2020distributionally, Conic_reformulation_KL_DRO_kocuk2020conic, DRQL_liu2022distributionally}.  If we choose $\phi(\tau)=\tau\ln(\tau)$,  we get \kldro with a KL-divergence-based ambiguity set. 
A closed-form solution to the inner maximization  \eqref{eqn:DRO} of \kldro is
 provided in \cite{KL-DRO_hu2013kullback}.
Nevertheless, the definition of $\phi$-divergence requires any distribution $P$ in the ambiguity set to be absolutely continuous with respect to the training distribution $P_0$ (denoted as $P<<P_0$), which means for any measurable set $\mathcal{A}$, $P_0(\mathcal{A})=0$ implies $P(\mathcal{A})=0$. This implicit restriction narrows the ambiguity set and consequently limits the robustness of DRO when facing test scenarios with support shift.

 The intrinsic difficulty of modeling ambiguity sets in DR-DFL stems from the infinite dimension of the probability space. This motivates us to model the ambiguity set in a parameterized space. Therefore, unlike these traditional ambiguity modeling, 
 we leverage diffusion models to directly learn the worst-case distribution in the context of DR-DFL exploiting their strong distribution modeling capability \cite{score-based_diffusion_SDE_song2021maximum,score_matching_song2019generative,NEURIPS2020_4c5bcfec,DDIM_song2020denoising} introduced as follows.

\subsection{Distribution Modeling via Diffusion Models}
The diffusion models learn the underlining distribution from a finite dataset and can generate more samples from the underlining distribution. They rely on forward and backward stochastic processes introduced as follows.

\textbf{Forward Process.} 
The forward process incrementally injects noise into the data, generating a sequence of perturbed samples. It begins with an initial sample $x_0\in\mathcal{R}^d$ drawn from the underlining distribution $P_0$, and evolves according to a stochastic process as:
\begin{equation}\label{eqn:forward}
\mathrm{d}x = k(x,t) \mathrm{d}t + g(t) \mathrm{d}w,
\end{equation}
where $k(\cdot,t): \mathcal{R}^d\rightarrow \mathcal{R}^d$ is a vector-valued function, $g(t)\in\mathcal{R}$, $w$ is a standard Wiener process and $\mathrm{d}w$ is white Gaussian noise. By the forward process, we get a collection of variables \(\{x_t\}_{t\in [0,T]}\). We use $P_t$ to represent the distribution of $x_t$ and $P_{t\mid 0}$ to denote the conditional distribution of $x_t$ given $x_0\sim P_0$. 

With a sufficiently long time \(T\), the marginal distribution $P_T(x_T)$ approximates a tractable prior distribution $\pi(x)$ which is typically chosen as a standard Gaussian distribution.

\textbf{Reverse Process.} A reverse diffusion process is associated with the forward equation in \eqref{eqn:forward} and is expressed as 
\begin{equation}\label{eqn:backward}
\mathrm{d}x = \left( k(x,t) -g(t)^2 \nabla_x \log P_t(x) \right) \mathrm{d}t + g(t)\mathrm{d}\bar{w},
\end{equation}
where $\bar{w}$ is a standard Wiener process in the reverse-time direction, $\nabla_x \log P_t(x)$ is the time-dependent score function.

\textbf{Score Matching.}
In the reverse process, the score function \(\nabla_x \log P_t(x)\) plays a critical role in directing the dynamics. To estimate the score function \(\nabla_x \log P_t(x)\), we train a score-based model \(s_\theta(x, t)\) based on samples generated from the forward diffusion process. The score-based model should minimize the following score-matching loss:
\[
J_{\mathrm{SM}}(\theta) =\int_0^T \mathbb{E}_{P_t(x)} \left[\lambda(t)\left\| \nabla_{x} \log P_{t}(x) - s_{\theta}(x, t) \right\|^2 \right] \mathrm{d}t,
\]
where $\lambda(t)>0$ is a positive weighting function. We usually approximate the score-matching loss by a tractable denoising score-matching loss up to a constant that does not rely on $\theta$:
\begin{equation}\label{eqn:score_match}
J\!(\theta) \!\!=\!\!\!\int_0^T \!\!\!\!\!\mathbb{E}_{P_0(x)P_{t\mid 0}(x'\mid x)} \!\!\left[\lambda(t)\!\!\left\| \nabla_{x'} \!\log P_{t\mid 0}\!(x'\!\!\mid \!\!x) \!- \!s_{\theta}(x, t) \right\|^2 \!\right] \!\!\mathrm{d}t,
\end{equation}

\textbf{Sampling.}
If we discretize the reverse process, initialize $x_T\sim \pi$ and replace $\nabla_x \log P_t(x)$ with the score-based model $s_{\theta}(x,t)$, we can generate samples with a Markov chain with $T$ steps:
\begin{equation}
x_{t-1}=x_t + [k(x_t,t)-g^2(t)s_{\theta}(x_t,t)]\Delta t +g(t)\sqrt{|\Delta t|}z_t,
\end{equation}
where $\Delta_t$ is a small enough constant and $z_t\sim\mathcal{N}(0,\mathbf{I})$. Most existing diffusion models generate samples following the Markov chain \cite{NEURIPS2020_4c5bcfec,score_matching_song2019generative,Diffusion_beat_GANs_dhariwal2021diffusion,DDIM_song2020denoising} and a common expression for the joint distribution of the reverse outputs is
\begin{equation}\label{eqn:probability_diffusion}
    P_{\theta}(x_{0:T})=\pi(x_T)\prod_{t=1}^TP_{\theta}(x_{t-1}\mid x_t),
\end{equation}
where \(P_{\theta}(x_{t-1}\mid x_t)=\mathcal{N}(x_{t-1};\mu_{\theta}(x_t,t),\Sigma_{\theta}(x_t,t))\). 

The following lemma shows that constraint on the denoising score-matching loss \eqref{eqn:score_match} implies that the constraint on the KL-divergence between $P_0$ and $P_{\theta}$ is satisfied. 
\begin{lemma}[Theorem 1 and Corollary 1 in \cite{score-based_diffusion_SDE_song2021maximum}]\label{KL_bound_diffusion}
Given the assumptions listed in Appendix A of \cite{score-based_diffusion_SDE_song2021maximum} \footnote{The assumptions require that the expected squared norm over $P_0$ and $\pi$ are bounded by any finite value, the functions $k(\cdot, t)$, $\nabla_x\log P_t(x)$, and $s_{\theta}(\cdot, t)$ are  Lipschitz continuous and upper bounded by a value related to $\|x\|$, $g(t)$ is a non-zero function, and 
$\int_{t=0}^T\int_{\mathcal{O}}\|P_t(x)\|_2^2+dg(t)^2\|\nabla_xP_t(x)\|_2^2\mathrm{d}x\mathrm{d}t$ for any open bounded set $\mathcal{O}$ and $\mathbb{E}\left[\exp(\frac{1}{2}\int_{t=0}^T\|\nabla_x\log P_t(x)-s_{\theta}(x,t)\|^2_2\mathrm{d}t)\right]$ are bounded by any finite value.
}, 
if the denoising score-matching loss \eqref{eqn:score_match} satisfies $J(\theta)\leq \epsilon$, the output distribution of the diffusion model $P_{\theta}(x_0)$ satisfies 
\[
D_{\mathrm{KL}}(P_0|| P_{\theta})\leq \epsilon + D_{\mathrm{KL}}(P_T||\pi)+C_1,
\]
where $P_T$ is the final-step output distribution of the forward process and $P_T\approx \pi$ by the design of diffusion models, and $C_1$ is a constant that does not rely on $\theta$.
\end{lemma}

Note that the KL-divergence $D_{\mathrm{KL}}(P_0|| P_{\theta})$ in Lemma \ref{KL_bound_diffusion} is not the KL-divergence $D_{\mathrm{KL}}(P_{\theta}||P_0)$ commonly used in \kldro. The former KL-divergence allows $P_{\theta}$ to have broader support space than $P_0$ ($P_0<<P$).   

\subsection{Diffusion-Based Ambiguity Set}
Lemma \ref{KL_bound_diffusion} implies that if we find a diffusion model distribution $P_{\theta}$ whose parameters satisfy $J(\theta)\leq \epsilon$, then $P_{\theta}$ stays close to the training distribution $P_{0}$ through a KL-divergence $D_{\mathrm{KL}}(P_0|| P_{\theta})$ up to a budget related to $\epsilon$. This KL-divergence is the reversed KL-divergence and allows $P_{\theta}$ to have broader support space than $P_0$. Therefore, we can define a parameterized ambiguity set based on the diffusion models without the support shift issue. 
The DR-DFL with diffusion-based ambiguity set is expressed as below. 
\begin{equation}\label{eqn:objective}
\min_{w\in\mathcal{W}}\max_{\theta\in\Theta} \mathbb{E}_{P_{\theta}(x)}[f(y^*(h_w(v)),c)], \quad \mathrm{s.t.}\;  J(\theta,S_0)\leq \epsilon,
\end{equation}
where $P_{\theta}(x), x=(v,c)$ is the output distribution of the diffusion model, $J(\theta,S_0)$ is the denoising score-matching loss of a diffusion model based on a training dataset $S_0$.

The diffusion-based ambiguity set leverages the powerful distribution modeling capabilities of diffusion models to enhance the generalization performance of DR-DFL. Specifically, diffusion models can generate diverse samples beyond the support of the training distribution, enabling the discovery of distributions with the worst decision-making performance, thereby yielding robust solutions. By constraining on the score-matching loss, the distributions in the ambiguity set remains consistent with the training data, striking a balance between average-case and worst-case performance. Moreover, the inner maximization in \eqref{eqn:objective} is conducted over a finite parameterized space rather than an infinite probability space, making the inner maximization tractable.

\section{\ouralg Algorithm}
Despite the advantages of diffusion-based ambiguity set, it is challenging to solve DR-DFL in \eqref{eqn:objective} due to the complexity of diffusion models.  We propose \ouralg algorithm in this section to solve this challenge. 
\subsection{Inner Maximization of \ouralg}
\begin{algorithm}[t]
\caption{Inner Maximization of \ouralg (\dagalg)}
\label{DAG}
\begin{algorithmic}[1]
\Require Training dataset $S_0$; ML model $h_w$, Adversary budget $\epsilon>0$; Step size $\eta>0$.
\State \textbf{Initialization}. Initialize the diffusion parameter $\theta$ and Lagrangian weight $\alpha>0$.
 \For{$k=1,2,\cdots, K$}
    \State Update diffusion model $\theta_k$ by solving \eqref{eqn:lagrangian_objective} given $\alpha$.  
    % $\theta_k= \max_{\theta} \mathbb{E}_{x\sim P_{\theta}}[l(h_w,x)] - \lambda \cdot  J_{SM}(\theta,S_0)$.
    \State Update the Lagrangian parameter $\alpha$: $\alpha\leftarrow \max\{\alpha
    +\eta (J(\theta_k,S_0)-\epsilon),0\}$.
 \EndFor
\State \Return Adversarial diffusion model $\theta_K$.
\end{algorithmic}
\end{algorithm}
Solving the inner maximization of \eqref{eqn:objective} presents two key challenges. First, the problem is a constrained non-convex optimization, making it difficult to maximize the objective and minimize the constraint violation simultaneously. Second, the objective function depends on the diffusion parameter $\theta$ through the probability function inside the expectation, which is computationally expensive to evaluate and differentiate.

We propose Algorithm \ref{DAG} to tackle the challenges of inner maximization. Observing that the constraint in \eqref{eqn:objective} is a budget constraint ($J(\theta)\geq 0$ and $\epsilon>0$), we can solve the constrained optimization by a dual learning method \cite{DMD_balseiro2020dual,lobos2021joint}. Algorithm \ref{DAG} adaptively learns a Lagrangian dual $\alpha>0$ to convert the inner constrained maximization into multi-step unconstrained optimizations below
\begin{equation}\label{eqn:lagrangian_objective}
\max_{\theta} \mathbb{E}_{P_{\theta}}[f(y^*(h_w(v)),c)]-\alpha J(\theta, S_0).
\end{equation}
We update $\alpha$ by dual gradient descent based on the denoising score-matching loss $J$: Increase $\alpha$ to emphasize more on the constraint satisfaction if  $J$ violates the budget and decrease $\alpha$ otherwise. After enough iterations, the dual variable converges to a near-optimal one that balances the objective maximization and constraint violation. 

Next, we transform the objective in \eqref{eqn:lagrangian_objective} into a differentiable term. By the Markov chain in \eqref{eqn:probability_diffusion}, we can express the joint probability of denoising outputs as 
\begin{equation}\label{eqn:backward_density}
    P_{\theta}(x_{0:T})= C \cdot e^{-\frac{\|x_T\|^2}{2}}\cdot e^{-\sum_{t=1}^T\frac{\|x_{t-1}-\mu_{\theta}(x_t,t)\|^2}{2\sigma_t^2}},
\end{equation}
where $C$ is a nomalizing constant, and $\sigma_t^2$ is the variance of the reserve noise at step $t$. Then, we can exploit tricks in policy optimization algorithms to transform the expected objective.

By the trick in vanilla policy gradient \cite{policy_gradient_sutton1999policy}, we can derive the gradient of the expected objective in \eqref{eqn:lagrangian_objective} as
\begin{equation}\label{eqn:vpg}
\begin{split}
\nabla_{\theta}&\mathbb{E}_{P_{\theta}(x)}[f(y^*(h_w(v)),c)]=\\
&\mathbb{E}_{P_{\theta}(x_{0:T})}[
    \nabla_{\theta} \ln P_{\theta}(x_{0:T}) \cdot f(y^*(h_w(v)),c)],
    \end{split}
\end{equation}
where $\ln P_{\theta}(x_{0:T})=\sum_{t=1}^T[x_{t-1}-\mu_{\theta}(x_t,t)]^2+C_2$ where $C_2$ is a constant. We can empirically calculate the expected gradient in \eqref{eqn:vpg} based on a dataset sampled from the reverse process of the diffusion model $P_{\theta}$.

Proximal Policy Optimization (PPO) \cite{PPO_schulman2017proximal} is believed to have more stable performance than vanilla policy gradient.
By PPO, we can transform the expected objective in \eqref{eqn:lagrangian_objective} into a differentiable form as
\begin{equation}\label{eqn:ppo}
\begin{split}
\mathbb{E}_{P_{\theta}}&[f(y^*(h_w(v)),c)]=\mathbb{E}_{P_{\theta_0}}[
   \min(r_{\theta}f(y^*(h_w(v)),c)), \\&\mathrm{clip}(r_{\theta}(x_{0:T}),1-\kappa,1+\kappa)\cdot f(y^*(h_w(v)),c))],
\end{split}
\end{equation}
where $\mathrm{clip}$ is the clipping function in PPO \cite{PPO_schulman2017proximal} with the clipping parameter $\kappa\in(0,1)$, the probability ratio is $r_{\theta}(x_{0:T})=\frac{P_{\theta }(x_{0:T})}{P_{\theta_0}(x_{0:T})}=\exp\{-\sum_{t=1}^T(\frac{\|x_{t-1}-\mu_{\theta}(x_t,t)\|^2}{2\sigma_{t}^2}-\frac{\|x_{t-1}-\mu_{\theta_0}(x_t,t)\|^2}{2\sigma_{t}^2})\}$, and the reference model $P_{\theta_0}$ can be a pre-trained diffusion model on the training dataset $S_0$. Similar as \eqref{eqn:vpg}, we can empirically calculate expected objective based on a the dataset sampled from the reverse process of the diffusion model $P_{\theta_0}$.
To reduce the training complexity, we can fix the parameters of the first $T-T'$ steps and only optimize the last $T'$ steps of the reverse process by choosing $r_{\theta}(x_{0:T'})=\exp\{-\sum_{t=1}^{T'}(\frac{\|x_{t-1}-\mu_{\theta}(x_t,t)\|^2}{2\sigma_{t}^2}-\frac{\|x_{t-1}-\mu_{\theta_0}(x_t,t)\|^2}{2\sigma_{t}^2})\}$.

\subsection{Min-Max Solution of \ouralg}
Now we are ready to solve the min-max problem in \eqref{eqn:objective}. The min-max solution is extended from the algorithm of Gradient Descent with Max-Oracle (GDMO) in \cite{GDMO_jin2020local}. For nonconvex-nonconcave min-max optimization problems, GDMO is proved to guarantee an approximate stationary solution with the approximation error depending on the error of inner-maximization. 

The algorithm flow of \ouralg is provided in Algorithm \ref{DAG-DRL}. Following the GDMO framework, \ouralg first runs \dagalg in Algorithm \ref{DAG} to search for the adversarial diffusion model $P_{\theta}$ that maximizes the expected loss of the current ML model $h_w$ within the diffusion ambiguity set. Next, given the updated diffusion model $P_{\theta}$, we need to update the ML parameter $w$ to minimize the expected decision objective $\mathbb{E}_{P_{\theta}}[f(y^*(h_w(v)),c)]$. One choice is to perform the gradient descent on the PPO transformation in \eqref{eqn:ppo}.  However, in order to provide the ML model with more diverse samples, we generate an adversarial dataset $S_{\theta}$ by the diffusion model $P_{\theta}$ and directly approximate the expected decision objective $\mathbb{E}_{P_{\theta}}[f(y^*(h_w(v)),c)]$ by $S_{\theta}$. Next, we can perform a gradient descent on the decision-focused objective based on the adversarial dataset $S_{\theta}$. The gradient of the objective can be obtained by differentiable optimization layers \cite{DFL_survey_mandi2024decision}. Alternatively, we can apply zero-order optimization methods \cite{zeroth-order_opt_liu2020primer} to estimate the gradients.

\begin{algorithm}[t]
\caption{\ouralg Algorithm}
\label{DAG-DRL}
\begin{algorithmic}[1]
\Require Training dataset $S_0$; Adversary budget $\epsilon>0$.  
\State \textbf{Initialization}. 

Initialize the ML model parameter $w$. 
 \For {$\mathrm{epoch} = 1, 2,\cdots, E$}
    \State Run \dagalg($h_w$,$\epsilon$) in Algorithm \ref{DAG} to update the diffusion model parameter $\theta$.
    \State Generate adversarial dataset $S_{\theta}$ based on the diffusion model $P_{\theta}$. 
    \State Update the ML model parameter $w$ based on $S_{\theta}$.
 \EndFor
\State \Return ML model $h_w$.
\end{algorithmic}
\end{algorithm}

\section{Case Study}\label{sec:simulation}
In this section, we evaluate the performance of \ouralg based on a simulation study on resource management for LLM inference serving. We present the problem statement and setups, followed by the empirical comparison of \ouralg and baselines. 
\subsection{System Model} 
    We give the system model for the application of Cloud Resource Provisioning \cite{online_resource_provisioning_comden2019online} for LLM Inference. Our objective is to develop a robust LLM workload predictor that achieves a good trade-off between utility performance and energy costs across diverse workload patterns. At each time step $i\in [N]$, the LLM inference workload is $c_i$, measured as the total number of input and output tokens assuming the best achievable LLM performance. Since the exact workload $c_i$ is unknown at the beginning of step $i$, the operator assigns an LLM instance with capacity $a_i$ based on the predicted workload $\hat{c}_i$. 
    While real-world resource provisioning involves multiple dimensions---including CPU, GPU cores, and memory---we abstract these complexities by defining $a_i$ as the token-handling capacity of the allocated LLM instance at time $i$. In other words, $a_i$ represents the number of tokens the instance can process in the slot $i$ while maintaining optimal LLM performance (no output length limit). 
    % This abstraction enables a tractable and focused modeling framework.
    
    To jointly capture the inference performance and the corresponding energy costs, we define an objective function that depends on the workload $c_i$ and the allocated capacity $a_i$. Specifically, the utility of assigning an LLM instance with capacity $a_i$ to process a workload of size $c_i$ is quantified by $\utility(a_i, c_i)$, which reflects the service quality achieved under the resource allocation $a_i$.
    When the allocated capacity fully satisfies the incoming workload ($a_i \geq c_i$), all requests can be processed with the highest performance, resulting in the maximum average utility $s(1)$. In contrast, when the allocated capacity is insufficient ($a_i < c_i$), the platform may either reject non-critical requests or restrict the output lengths of LLMs, leading to degraded service quality \cite{LLM_inference_serving_survey_li2024llm,yang2024queueing}. In such cases, the average utility is modeled as $s\left(\frac{a_i}{c_i}\right)$, where $s(\cdot)$ is an increasing function of the ratio of the allocated capacity to the demanded capacity.
    The overall $\utility$ model is defined as:
    \begin{equation}\label{eqn:utility}
    \utility (a_i, c_i) = 
        \begin{cases}
        s(1) \cdot c_i, & \text{if } a_i \geq c_i, \\
        s\left(\frac{a_i}{c_i}\right) \cdot c_i, & \text{if } a_i < c_i.
        \end{cases}
    \end{equation}

    The function $\cost(a_i, c_i)$ captures the total energy cost associated with assigning an LLM instance with capacity $a_i$ for a time slot $i$ with an inference workload $c_i$. The instance processes the workload of $\min(a_i, c_i)$ using activated computing resources, incurring energy consumption of $\pact \cdot \min(a_i, c_i)$. If the allocated capacity exceeds the actually demanded workload ($a_i > c_i$), the overly allocated capacity $(a_i - c_i)^+$ incurs additional energy consumption $\pidle \cdot (a_i - c_i)^+$ at a lower power rate , due to power-saving techniques such as GPU frequency scaling. 
    The total energy cost is scaled by Power Usage Effectiveness (PUE) $\omega$, and is modeled as:
    \begin{equation}
    \cost (a_i, c_i) = \omega \cdot \Big(\pact \cdot \min(a_i, c_i) + \pidle \cdot (a_i - c_i)^+ \Big)
    \end{equation}

    The overall objective of capacity provisioning is to maximize the net utility of LLM inference serving, accounting for both service performance and energy cost. The optimization problem is formulated as:
    \begin{equation}
    \begin{split}
    \max_{\forall i, a_i \in [a_{\min}, a_{\max}]} &\mathcal{R}(a_{1:N},c_{1:N})\coloneqq \\
    & \sum_{i=1}^N \Big[\utility(a_i, c_i) - \gamma \cdot \cost(a_i, c_i)\Big],
    \end{split}
    \end{equation}    
    where $[a_{\min}, a_{\max}]$ denote the range of allowable LLM processing capacities, and $\gamma>0$ is a scaling coefficient that unifies the units of $\utility$ and $\cost$.

\subsection{Experiment Setups}
    \subsubsection{System Setups}
       In the LLM inference serving system, the capacity for an LLM instance is decided at the beginning of each time slot and remains constant within the time slot to avoid unstable service quality. A sequence example includes consecutive $N=28$ time slots.  In the simulation,  we set the maximum processing speed of an LLM instance with multiple GPUs as $4\times 10^5$ tokens per time slot.  
       In addition, we set the power consumption per token as $P_{\mathrm{act}}=4\times 10^{-6} \mathrm{kWh}$ based on the estimation of GPT-3 \cite{GPT-3_brown2020language} that GPT-3 consumes an order of 0.4 kWh of electricity to generate 100 pages of content.
  The idle power consumption is set as $P_{\mathrm{idle}}=1.4\times 10^{-6} \mathrm{kWh}$ which is about one third of the activated power consumption.  The PUE is set as $\omega=1.1$. 
        
        For the utility model in \eqref{eqn:utility}, we adopt a logarithmic function $s(X) = B \log(AX + 1)$, which captures the diminishing returns of resource allocation. Logarithmic utility functions are widely used in network economics and resource allocation. Similar logarithmic utility functions have also been employed by Low \emph{et al.}~\cite{duality_model_2003} to model the utility of network flows in TCP congestion control. Moreover, Stephen \emph{et al.}~\cite{boyd2004convex} highlight that such utility functions possess desirable properties---monotonicity and strict concavity---which make them well-suited for modeling fair resource allocation problems. This form can be readily substituted with alternative utility functions that reflect revenue or service quality in practical LLM serving scenarios.
        We choose the parameters in the utility function as, $A = 20$, and $B = 0.2$.  The cost coefficient $\mu$ is set as $0.34$
        % $0.05$.

    \subsubsection{Baselines}
        The baselines which are compared with our algorithms in our experiments are introduced as below.

        \textbf{Decision-Focused Learning} (\ml): This method \cite{DFL_survey_mandi2024decision} trains the ML to optimize the decision objective without considering distributionally robustness. 
        
        \textbf{Wasserstein-based DRO} (\wdro): This is a \dro  algorithm where the ambiguity set is defined by the Wasserstein measure. In the experiments, we choose the FWDRO algorithm in \cite{staib2017distributionally} which applies to general objectives, and replace its loss function with our decision-focused objective.  
        
   \textbf{KL-divergence-based DRO} (\kldro): This is a \dro algorithm where the ambiguity set is defined by the KL divergence. We choose the commonly-used \kldro solution derived in \citecitenew{KL-DRO_hu2013kullback} and replace its loss function with our decision-focused objective.
        
        \textbf{Data Augmentation} (\da): Data augmentation techniques are commonly used to improve the generalization performance of ML by incorporating more diverse training samples \cite{Data_augmentation_laskin2020reinforcement}. In the experiments, we inject new samples to the training datasets by adding Gaussian, Perlin or Cutout noise. 
    
    \subsubsection{Datasets}
        The experiments are conduct based on the dataset of Azure LLM Inference Traces \cite{patel2024splitwise,stojkovic2024dynamollmdesigningllminference}. 
        The dataset captures time series of input and output token counts for each service request in the years 2023 and 2024 from two production-grade LLM inference services deployed within Azure, targeting code-related and conversational tasks, respectively. 
    To assess the generalization performance in different testing distributions, we split the datasets into a training dataset and several testing datasets with different distribution shifts. 
        
        All ML models are trained on the 2023-Conversations (\texttt{23V} (Train)) dataset with 751 sequence samples and evaluated on different testing datasets with dataset sizes ranging from 798 to 4320. The distributional discrepancy between each testing set and the training set is quantified by Wasserstein distance shown under the dataset names in Table \ref{main_table}.  The testing sets are listed as below with their time, LLM task and acronym: 2023-Conversations (\texttt{23V} (Test)),  2023-Code (\texttt{23D}), 2024-Code (\texttt{24D}), and 2024-Conversations (\texttt{24V}). To increase the diversity of the testing sets, we merge instances from two original datasets in an half-to-half way and get three additional testing sets: 2023-Code\&2024-Code (\texttt{23D24D}), 2024-Code\&2024-Conversations (\texttt{24D24V}) and 2023-Code\&2024-Conversations (\texttt{23D24V}).
    
    \subsubsection{Training Setups} 
        The experimental setup is divided into the following parts:
        
        \textbf{Predictor}: The workload predictors in \ouralg and all the baselines share the same two-layer stacked LSTM architecture with 128 and 64 hidden neurons.
        
        \textbf{Diffusion Model}: The diffusion model in \ouralg is $\DDPM$ \citecitenew{NEURIPS2020_4c5bcfec} with U-Net backbone which has $T = 500$ steps in a forward or a backward process. 
        
        \textbf{Training}: For \ouralg, we adopt the PPO-based  reformulation in \eqref{eqn:ppo} for inner maximization. We train the reference $\DDPM$ $\theta_0$ in \eqref{eqn:ppo} based on the original training dataset \texttt{23V} (Train) and use it to generate an initial dataset $\zinit$ to calculate $r_{\theta}$ in \eqref{eqn:ppo}. The sampling variance of $\DDPM$ is chosen from a range $[0.05, 0.1]$. To improve training efficiency, only the last $T'=10$ backward steps of the $\DDPM$ model are fine-tuned by \eqref{eqn:ppo}. We choose a slightly higher clipping parameter $\kappa=0.4$ in \eqref{eqn:ppo} to encourage the maximization while maintaining stability. We choose $\epsilon=0.03$ as \dagalg's adversarial budget which gives the best average performance over all validation datasets. We choose $\eta = 0.01$ as the rate to update the Lagrangian parameter $\alpha$ in Algorithm \ref{DAG}. We use the Adam optimizer with a learning rate of $10^{-6}$ for both the diffusion training in the maximization and the predictor update in minimization. The diffusion model is trained for 10 inner epochs with a batch size of 64.  The predictor is trained for 15 epochs with a batch size of 64.
        
        For the baseline methods, we choose the same neural network architecture as \ouralg. We carefully tuned the hyperparameters of the baseline algorithms to achieve optimal average performance over all validation datasets. 
        For \wdro, we consider the Wasserstein distance with respect to $l_2-$norm and set the adversarial budget as $\epsilon = 2$. For \kldro, we choose the adversarial budget $\epsilon = 2$.
        The predictors in both baseline \dro methods are trained by Adam optimizer with a learning rate of $2 \times 10^{-5}$. Both baselines are trained for 100 epochs with a batch size of 64.
    
\subsection{Experiment Results}
          \subsubsection{Default Setting}
        \begin{table}[htbp]
\centering
\caption{Test regret on different datasets.}
\renewcommand{\arraystretch}{1.2}
\resizebox{0.49\textwidth}{!}{%
\begin{tabular}{lcccccc}
\toprule
\multicolumn{1}{c}{\multirow{2}{*}{\textbf{Dataset}}}   & \multicolumn{6}{c}{\textbf{Algorithms}} \\
\cmidrule(lr){2-7}
 & \textbf{\ouralg} & \textbf{\kldro} & \textbf{\wdro} & \textbf{Cutout} & \textbf{Gaussian}  & \textbf{\ml}\\
\midrule

\multicolumn{1}{c}{\makecell{\textbf{23V(Test)} \\ (0.0001)}}  & \textbf{0.0518} & 0.0797 & 0.0682 & 0.0967 & 0.0949  & 0.1298\\
\multicolumn{1}{c}{\makecell{\textbf{24V} \\ (0.0961)}}   & \textbf{0.0283} & 0.0604 & 0.0828 & 0.0707 & 0.0716  & 0.1003\\
\multicolumn{1}{c}{\makecell{\textbf{23D} \\ (0.1459)}} & \textbf{0.2213} & 0.2967 & 0.3087 & 0.3241 & 0.3304  & 0.3993\\
\multicolumn{1}{c}{\makecell{\textbf{24D} \\ (0.1011)}} & \textbf{0.2775} & 0.4558 & 0.6266 & 0.4648 & 0.5004  &0.6103\\
\multicolumn{1}{c}{\makecell{\textbf{23D24D} \\ (0.1565)}} & \textbf{0.3140} & 0.5071 & 0.6894 & 0.5215 & 0.5643  & 0.6828\\
\multicolumn{1}{c}{\makecell{\textbf{23D24V} \\ (0.1357)}} & \textbf{0.0703} & 0.1214 & 0.1680 & 0.1317 & 0.1376  & 0.1791\\
\multicolumn{1}{c}{\makecell{\textbf{24D24V} \\ (0.0687)}} & \textbf{0.1814} & 0.3072 & 0.3976 & 0.3259 & 0.3471  &0.4360\\
\midrule
\multicolumn{1}{c}{\textbf{Average}} & \textbf{0.1635} & 0.2612 & 0.3345 & 0.2765 & 0.2923  &  0.3625\\
\midrule
\multicolumn{1}{c}{\textbf{Worst}} & \textbf{0.3140} & 0.5071 & 0.6894 & 0.5215 & 0.5643 &  0.6828\\
\bottomrule
\end{tabular}
}
\label{main_table}
\end{table}

        We give a comprehensive comparison between \ouralg and different decision-focused baseline algorithms on various testing datasets with different distribution shifts.
  We evaluate the performance by the normalized \(\textit{Regret}(\mathrm{alg}) = (\bar{\mathcal{R}}_{\mathrm{opt}}-\bar{\mathcal{R}}_{\mathrm{alg}})/\bar{\mathcal{R}}_{\mathrm{opt}} \), where $\bar{\mathcal{R}}_{\mathrm{alg}}$ denotes the mean performance of the algorithm on a testing dataset and $\bar{\mathcal{R}}_{\mathrm{opt}}$ represents the optimal mean performance on the same testing dataset. The regrets and their average and maximum values over all testing datasets are shown in Table~\ref{main_table}, We can find that \ouralg outperforms all \dro algorithms across all datasets. Specifically, while \kldro and \wdro also achieve notable improvements over the standard \ml method, \ouralg attains an average regret of 0.1635 over all datasets, exceeding the average performance of \kldro and \wdro by 37.4\% and 51.1\%, respectively. Furthermore, on the \textbf{23D24D} dataset with the largest distribution shift, all baselines reach the maximum regret, but \ouralg achieves a maximum regret of 0.3140, surpassing \kldro and \wdro by 38.0\% and 54.4\%, respectively. The advantages of \ouralg come from the use of diffusion model to construct the ambiguity set. Compared to the ambiguity sets with KL divergence, diffusion-based ambiguity set allows \ouralg to generate diverse samples beyond the support of the training distribution, discover distributions that lead to the worst-case decision-making performance. Meanwhile, by restricting the denoising score matching loss by a budget $\epsilon$, \ouralg ensures that the distributions in the ambiguity set are consistent with the underlining data distribution. This avoids the overly-broad relaxation of ambiguity set in the \wdro solution and achieve performance improvements far exceeding \wdro. The comparison with DRO baselines demonstrate that \ouralg with the diffusion-based ambiguity set is superior in achieving a favorable performance balance between average and worst-case testing environments.
  
    In addition, in Table~\ref{main_table}, we compare \ouralg with two common data augmentation methods which inject new samples by adding \textit{Cutout} or \textit{Gaussian} noise. 
    For \da with \textit{Cutout} noise, each training data point at each time step is masked to zero with a probability of 5\%, and the cutout dataset is added to original training dataset. 
    For \da with \textit{Gaussian} noise, a standard Gaussian noise scaled by 5\% of the maximum value of the original data point is added to each data point. at each time step. Then, the decision-focused training is applied on the augmented training datasets. As shown in Table~\ref{main_table}, both data augmentation methods provide obvious performance improvements for both average and maximum performance compared to \ml since \da makes the training data more diverse and so enhances the generalization performance. \textit{Cutout} augmentation performs better than Gaussian augmentation, achieving an average Regret of 0.2765, which represents a 23.7\% improvement over \ml. However, \da methods have a limited performance improvement especially for the maximum regret. This is because \da methods do not optimize for the worst-case performance and they cannot effectively inject samples that are important for the decision objective.  By contrast, \ouralg effectively optimizes for the worst-case and decision aware objectives, thus significantly outperforming \da methods on both average and worst-case performance. 

\begin{figure*}[htbp]
        \centering
        \begin{minipage}[b]{0.33\textwidth}
            \centering
            \includegraphics[width=\textwidth]{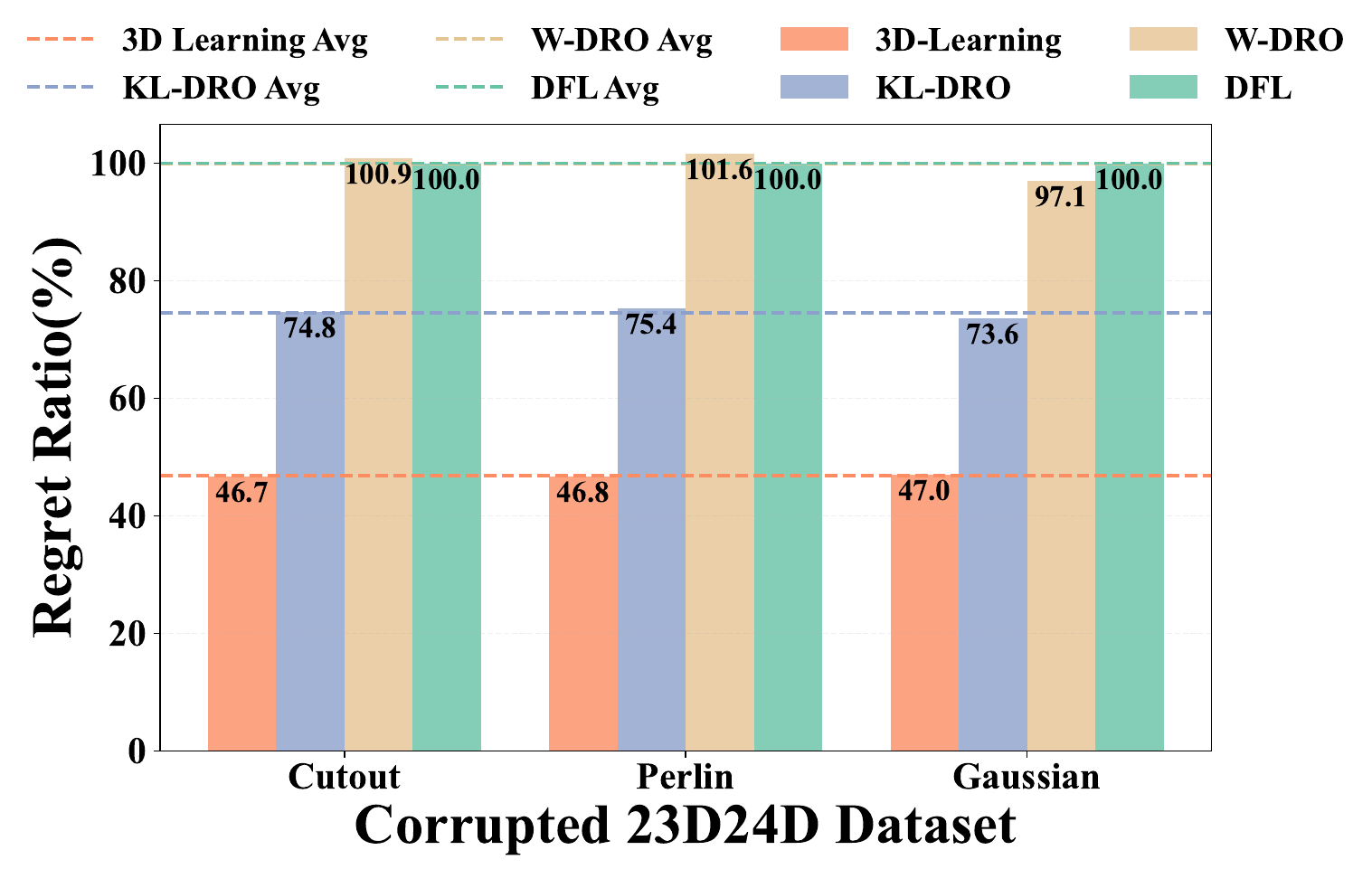}
            \caption{Robustness evaluation under diverse noisy corruptions on $\textbf{23D24D}$ dataset.}
            \label{Corruption Robustness Comparison}
        \end{minipage}
        \hfill
        \begin{minipage}[b]{0.33\textwidth}
            \centering
            \includegraphics[width=\textwidth]{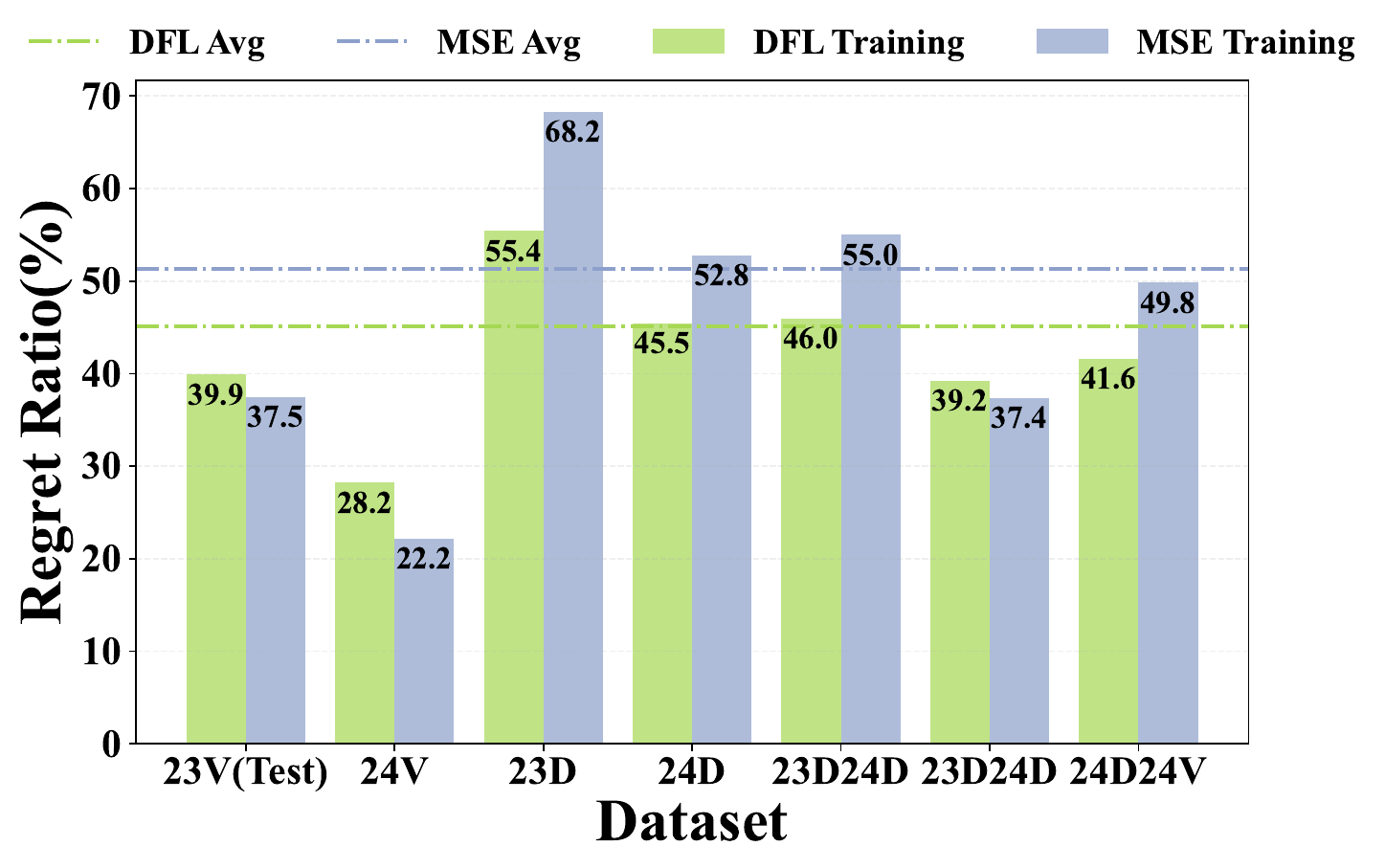}
            \caption{\ouralg with decision-focused training and MSE training.}
            \label{DFL-MSE Comparison}
        \end{minipage}
        \hfill
        \begin{minipage}[b]{0.32\textwidth}
            \centering
            \includegraphics[width=\textwidth]{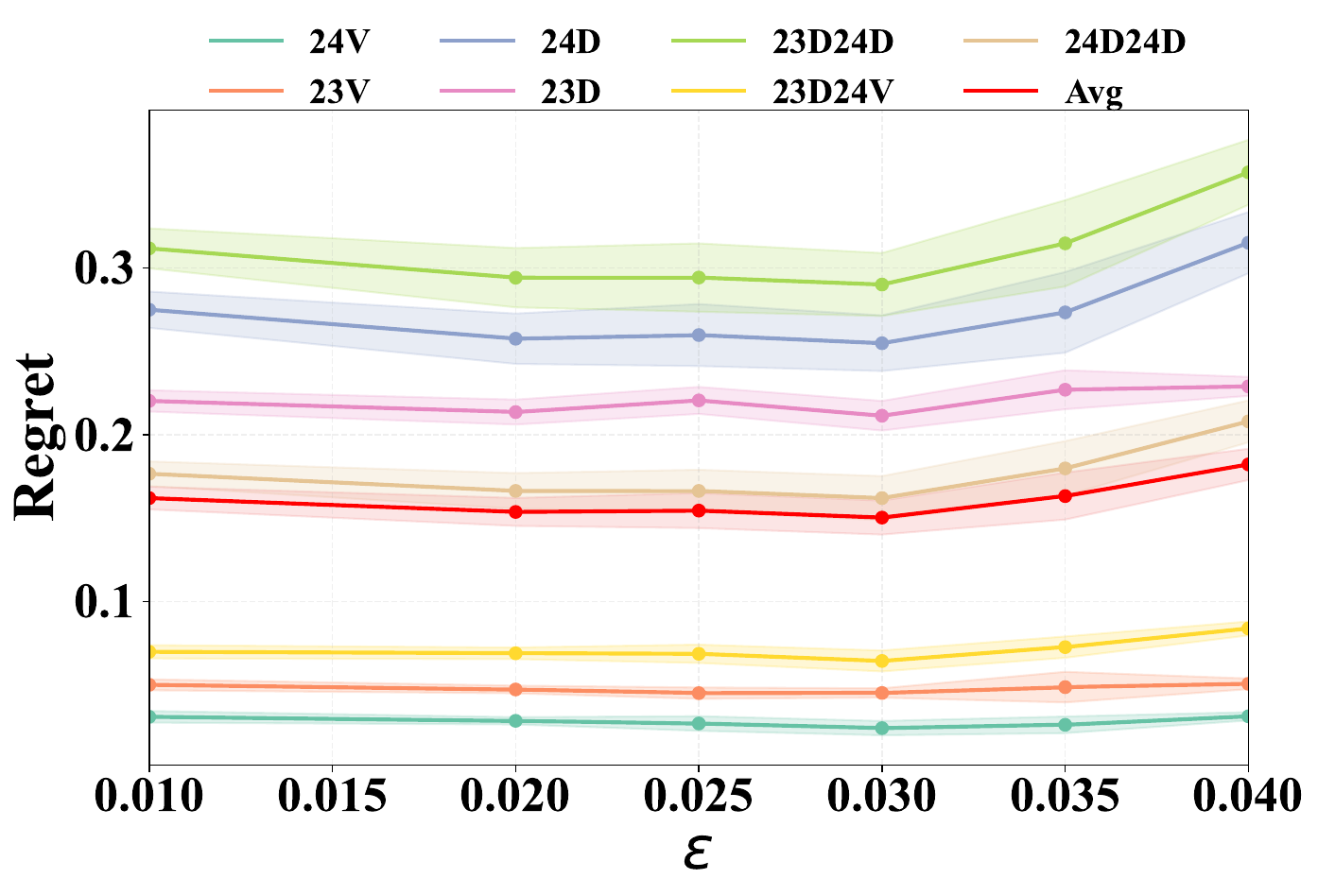}
            \caption{Effect of budget $\epsilon$ on \ouralg performance.}
            \label{budget_selection}
        \end{minipage}
    \end{figure*}
    \subsubsection{Evaluation on Corrupted Datasets}
       In Fig. \ref{DFL-MSE Comparison}, we access the performance of distributionally robust algorithms on corrupted testing environments.  We create corrupted testing datasets by injecting Cutout, Perlin, or Gaussian noise to the original testing dataset (\texttt{23D24D}) which has the largest Wasserstein distribution discrepancy compared to the training dataset.  For Cutout noise, each per-round data point is masked to zero with a probability of 0.5\%. The added Perlin noise has a magnitude of 5\% of the maximum value of the original data point. Standard Gaussian noise scaled by 10\% of the original maximum value is added to the original dataset. The regret ratios ($\textit{Regret}(\mathrm{alg})/\textit{Regret}(\ml)\times 100$ \%) with the regret of \ml as the baseline are illustrated in Fig~\ref{Corruption Robustness Comparison}. Under various noisy perturbations, \ouralg achieves an average Regret Ratio of only 46.8\% of \ml, while significantly outperforming \kldro and \wdro by 27.7\% and 53.0\%, respectively. These results demonstrate that by diffusion augmented DRO, \ouralg exhibits outstanding robustness against noisy corruptions compared to existing methods.
    
    \subsubsection{Effects of Training Objectives}
        Next, we investigate the effects of training objectives (decision-focused or MSE training) on \ouralg. The MSE training replaces the decision objective in \ouralg with the mean squared prediction error.  The regret ratios of both training strategies are illustrated in Fig~\ref{DFL-MSE Comparison}. Decision-focused training achieves, on average, a 6.1\% improvement over MSE training in decision performance across various datasets. However, decision-focused training does not win for all datasets: it performs slightly worse on the datasets \texttt{23V} (Test) \texttt{24V} and \texttt{23D24D}. This is because distributionally robust learning optimizes for the worst-case distribution but not for every distinct distribution. 
Nevertheless, by the end-to-end training strategy, decision-focused training demonstrates superior performance in most cases compared to prediction-focused training. 
    
    \subsubsection{Effects of \dro Budget}
        We investigate the effects of the critical budget parameter $\epsilon$ in \eqref{eqn:objective} on the performance of \ouralg. 
        As shown in Fig~\ref{budget_selection}, the regret-$\epsilon$ curves across all datasets exhibit a concave shape, achieving the optimal average performance at around $\epsilon = 0.03$. When $\epsilon$ falls below this threshold and continues to decrease, the diffusion-modeled distribution becomes overly constrained to the training data, limiting the ability of \ouralg to generalize to OOD datasets. Conversely, when $\epsilon$ exceeds this threshold and continues to increase, \ouralg with an overly large ambiguity set can conservatively over-optimize the decision objective on irrelevant distributions, which can result in degraded performance on real OOD datasets. Therefore, it is critical to select an appropriate budget $\epsilon$ to generate effective adversarial distributions, striking a desirable trade-off between average and worst-case performance.

    \subsubsection{Effects of Diffusion Model Setups}
    \begin{table}[htbp]
\centering
\caption{Test Regret of \ouralg with different Diffusion Models and finetuning backward steps $T'$.}
\renewcommand{\arraystretch}{1.2}
\resizebox{0.49\textwidth}{!}{%
\begin{tabular}{lcccccc}
\toprule
\multicolumn{1}{c}{\multirow{1}{*}{\textbf{Regret}}}

 & \textbf{DFU-128} & \textbf{DFU-96} & \textbf{DFU-64} & \textbf{$T'$-15} & \textbf{$T'$-10}  & \textbf{$T'$-5}\\
\midrule

\multicolumn{1}{c}{\textbf{Average}} & 0.1635 & 0.1699 & 0.1721 & 0.1547 & 0.1635  &  0.1891\\
\midrule
\multicolumn{1}{c}{\textbf{Worst}} & 0.3140 & 0.3300 & 0.3304 & 0.3031 & 0.3140 &  0.3815\\
\bottomrule
\end{tabular}
}
\label{size_table}
\end{table}

The regret of diffusion model setups of \ouralg is given in Table~\ref{size_table}. \textbf{DFU-$M$} means the diffusion model with 4-level UNet backbone which has $M$ channels in the first level and $2M$ channels in the other levels. 
$T'-N$ means the last $N$ backward steps of \textbf{DFU-128} are finetuned for inner maximization.
As shown in Table~\ref{size_table}, evaluation across all test sets shows that a larger diffusion model can result in lower regrets because it has higher expressive capacity to represent the adversarial distributions.
In addition, fine-tuning more backward steps improves the performance because it provides more flexibility to solve the inner maximization.

    \subsubsection{Computational Overhead}
       We benchmark all methods on the above time-series task, measuring training runtime and GPU memory on a single NVIDIA RTX 6000 Ada GPU.
During training, \ml and \da have similar computational overhead (runtime $\approx$2100 s, GPU memory $<35$ MB), whereas \dro methods are substantially more expensive: \ouralg requires a runtime of 4900 s. In terms of memory consumption, \wdro and \kldro use over 400 MB of GPU memory, while \ouralg requires significantly more (6.8 GB) due to the usage of diffusion model.
        Importantly, this overhead is confined to training: at test time, all methods exhibit similar inference overhead (84 s, 418 MB), indicating no additional inference overhead for \ouralg.

\section{Related Works}

Our work is closely related to the literature on DRO. Most existing DRO algorithms construct ambiguity sets using either Wasserstein distances \cite{DRO_mohajerin2018data,DRL_chen2020distributionally,Wasserstein_DRO_gao2023distributionally,risk_stochastic_opt_zhao2018data}  or $\phi$-divergences \cite{KL-DRO_hu2013kullback, DR_BO_divergence_husain2023distributionally, DRBO_kirschner2020distributionally,chance-constrained-jiang2016data}. However, these approaches often involve optimization over an infinite-dimensional probability space, which poses significant computational challenges and can result in sub-optimal solutions. Recent studies \cite{DRL_environment_generation_ren2022distributionally,michelmodeling} represent the adversarial distributions within traditional  Wasserstein-based or KL divergence-based ambiguity sets.  
However, due to the reliance on these traditional ambiguity sets, the intractability issue or the support shift issue still exist. 
In contrast, \ouralg constructs a novel ambiguity set based on the score-matching loss of the diffusion model, which 
offers better flexibility in searching adversarial distributions.

Robust DFL is also studied in recent literature. 
Ma \emph{et al.}~\cite{ma2024differentiable} introduce a differentiable parameterized Second-Order Cone (SOC) to define the ambiguity set and propose an end-to-end framework that trains ML to predict the ambiguity set for the downstream DRO task.
In comparison, our DR-DFL framework is fundamentally different, as it addresses the distributionally robustness problem during the training of the context predictor, rather than focusing on inference-time DRO tasks as in \cite{ma2024differentiable}. Moreover, \ouralg constructs ambiguity sets using diffusion models, which capture diverse adversarial distributions, thereby offering greater expressiveness.
Wang \emph{et al.}~\cite{Gen-DFL_wang2025gen} proposed a Generate-then-Optimize framework that trains a diffusion model to generate data for downstream statistical optimization, targeting the conditional value-at-risk (CVaR) objective. While closely related, it primarily addresses risk mitigation under in-distribution scenarios, whereas \ouralg is designed to enhance robustness in out-of-distribution (OOD) environments.

\section{Conclusion}
Our work focuses on DR-DFL which models many critical PTO applications in networking. We propose a diffusion augmented algorithm \ouralg to improve the OOD generalization of DR-DFL. Specifically, by leveraging diffusion-based ambiguity modeling, \ouralg enables the search of worst-case distributions in the parameterized space of diffusion models, which achieves a good balance between average and worst-case performances. Extensive simulation results on resource provisioning for LLM inference confirm the effectiveness of \ouralg, demonstrating substantial performance gains and enhanced robustness under perturbations. These findings highlight the potential of diffusion models as a powerful tool for distributionally robust, decision-driven learning in dynamic and uncertain environments. For future work, the diffusion-based ambiguity set can be applied to other DRO problems.

\section*{Acknowledgment}
Jiaqi Wen and Jianyi Yang were supported in part by University of Houston Start-up Funds 74825 [R0512039] and 74833 [R0512042]. Lei Fan was supported by UH Energy Transition Institute.


\begin{thebibliography}{10}

\bibitem{DMD_balseiro2020dual}
Santiago Balseiro, Haihao Lu, and Vahab Mirrokni.
\newblock Dual mirror descent for online allocation problems.
\newblock In Hal~Daumé III and Aarti Singh, editors, {\em ICML}, volume 119 of
  {\em Proceedings of Machine Learning Research}, pages 613--628. PMLR, 13--18
  Jul 2020.

\bibitem{boyd2004convex}
Stephen~P Boyd and Lieven Vandenberghe.
\newblock {\em Convex Optimization}.
\newblock Cambridge University Press, 2004.

\bibitem{GPT-3_brown2020language}
Tom Brown, Benjamin Mann, et~al.
\newblock Language models are few-shot learners.
\newblock 33:1877--1901, 2020.

\bibitem{energy_measurement_AI_caspart2022precise}
René Caspart, Sebastian Ziegler, Arvid Weyrauch, et~al.
\newblock Precise energy consumption measurements of heterogeneous artificial
  intelligence workloads.
\newblock In {\em HPCC}, pages 108--121. Springer, 2022.

\bibitem{DRL_chen2020distributionally}
Ruidi Chen and Ioannis~Ch. Paschalidis.
\newblock Distributionally robust learning.
\newblock {\em Foundations and Trends® in Optimization}, 4(1-2):1--243, 2020.

\bibitem{AI-data-center-grid-colangelo2025turning}
Philip Colangelo, Ayse~K Coskun, Jack Megrue, et~al.
\newblock Turning ai data centers into grid-interactive assets: Results from a
  field demonstration in phoenix, arizona.
\newblock {\em arXiv preprint arXiv:2507.00909}, 2025.

\bibitem{online_resource_provisioning_comden2019online}
Joshua Comden, Sijie Yao, Niangjun Chen, Haipeng Xing, and Zhenhua Liu.
\newblock Online optimization in cloud resource provisioning: Predictions,
  regrets, and algorithms.
\newblock {\em SIGMETRICS}, 2019.

\bibitem{report_demand_surge_davenport2024ai}
Carly Davenport, CFA Singer, N~Mehta, B~Lee, and J~Mackay.
\newblock Ai data centers and the coming us power demand surge.
\newblock {\em PDF). Goldman Sachs. Archived from the original (PDF) on}, 26,
  2024.

\bibitem{Diffusion_beat_GANs_dhariwal2021diffusion}
Prafulla Dhariwal and Alexander Nichol.
\newblock Diffusion models beat gans on image synthesis.
\newblock 34:8780--8794, 2021.

\bibitem{end-to-end-learning_donti2017task}
Priya Donti, Brandon Amos, and J.~Zico Kolter.
\newblock Task-based end-to-end model learning in stochastic optimization.
\newblock 30, 2017.

\bibitem{Wasserstein_DRO_gao2023distributionally}
Rui Gao and Anton Kleywegt.
\newblock Distributionally robust stochastic optimization with wasserstein
  distance.
\newblock {\em Mathematics of Operations Research}, 48(2):603--655, 2023.

\bibitem{NEURIPS2020_4c5bcfec}
Jonathan Ho, Ajay Jain, and Pieter Abbeel.
\newblock Denoising diffusion probabilistic models.
\newblock In H.~Larochelle, M.~Ranzato, R.~Hadsell, M.F. Balcan, and H.~Lin,
  editors, {\em NeurIPS}, volume~33, pages 6840--6851. Curran Associates, Inc.,
  2020.

\bibitem{KL-DRO_hu2013kullback}
Zhaolin Hu and L~Jeff Hong.
\newblock Kullback-leibler divergence constrained distributionally robust
  optimization.
\newblock {\em Available at Optimization Online}, 1(2):9, 2013.

\bibitem{DR_BO_divergence_husain2023distributionally}
Hisham Husain, Vu~Nguyen, and Anton van~den Hengel.
\newblock Distributionally robust bayesian optimization with
  $\phi$-divergences.
\newblock 2023.

\bibitem{chance-constrained-jiang2016data}
Ruiwei Jiang and Yongpei Guan.
\newblock Data-driven chance constrained stochastic program.
\newblock {\em Mathematical Programming}, 158(1):291--327, 2016.

\bibitem{GDMO_jin2020local}
Chi Jin, Praneeth Netrapalli, and Michael Jordan.
\newblock What is local optimality in nonconvex-nonconcave minimax
  optimization?
\newblock In Hal~Daumé III and Aarti Singh, editors, {\em ICML}, volume 119 of
  {\em Proceedings of Machine Learning Research}, pages 4880--4889. PMLR,
  13--18 Jul 2020.

\bibitem{DRBO_kirschner2020distributionally}
Johannes Kirschner, Ilija Bogunovic, Stefanie Jegelka, and Andreas Krause.
\newblock Distributionally robust bayesian optimization.
\newblock In Silvia Chiappa and Roberto Calandra, editors, {\em Proceedings of
  the Twenty Third AISTATS}, volume 108 of {\em Proceedings of Machine Learning
  Research}, pages 2174--2184. PMLR, 26--28 Aug 2020.

\bibitem{Conic_reformulation_KL_DRO_kocuk2020conic}
Burak Kocuk.
\newblock Conic reformulations for kullback-leibler divergence constrained
  distributionally robust optimization and applications.
\newblock {\em IJOCTA}, 11(2):139–151, April 2021.

\bibitem{Wasserstein_DRO_kuhn2019wasserstein}
Daniel Kuhn, Peyman~Mohajerin Esfahani, Shafieezadeh Nguyen, et~al.
\newblock Wasserstein distributionally robust optimization: Theory and
  applications in machine learning.
\newblock pages 130--166. 2019.

\bibitem{Data_augmentation_laskin2020reinforcement}
Misha Laskin, Kimin Lee, Adam Stooke, Lerrel Pinto, Pieter Abbeel, and Aravind
  Srinivas.
\newblock Reinforcement learning with augmented data.
\newblock 33:19884--19895, 2020.

\bibitem{LLM_inference_serving_survey_li2024llm}
Baolin Li, Yankai Jiang, Vijay Gadepally, and Devesh Tiwari.
\newblock Llm inference serving: Survey of recent advances and opportunities.
\newblock In {\em 2024 IEEE HPEC}, pages 1--8, 2024.

\bibitem{zeroth-order_opt_liu2020primer}
Sijia Liu, Pin-Yu Chen, Bhavya Kailkhura, Gaoyuan Zhang, Alfred~O Hero~III, and
  Pramod~K Varshney.
\newblock A primer on zeroth-order optimization in signal processing and
  machine learning: Principals, recent advances, and applications.
\newblock {\em IEEE Signal Processing Magazine}, 37(5):43--54, 2020.

\bibitem{pricing_demand_response_liu2014pricing}
Zhenhua Liu, Iris Liu, Steven Low, and Adam Wierman.
\newblock Pricing data center demand response.
\newblock {\em ACM SIGMETRICS Performance Evaluation Review}, 42(1):111--123,
  2014.

\bibitem{DRQL_liu2022distributionally}
Zijian Liu, Qinxun Bai, Jose Blanchet, Perry Dong, Wei Xu, Zhengqing Zhou, and
  Zhengyuan Zhou.
\newblock Distributionally robust $ q $-learning.
\newblock In {\em ICML}, pages 13623--13643. PMLR, 2022.

\bibitem{lobos2021joint}
Alfonso Lobos, Paul Grigas, and Zheng Wen.
\newblock Joint online learning and decision-making via dual mirror descent.
\newblock In Marina Meila and Tong Zhang, editors, {\em ICML}, volume 139 of
  {\em Proceedings of Machine Learning Research}, pages 7080--7089. PMLR,
  18--24 Jul 2021.

\bibitem{duality_model_2003}
S.H. Low.
\newblock A duality model of tcp and queue management algorithms, 2003.

\bibitem{edge_survey_luo2021resource}
Quyuan Luo, Shihong Hu, Changle Li, Guanghui Li, and Weisong Shi.
\newblock Resource scheduling in edge computing: A survey.
\newblock {\em IEEE Communications Surveys and Tutorials}, 23(4):2131--2165,
  2021.

\bibitem{ma2024differentiable}
Xutao Ma, Chao Ning, and Wenli Du.
\newblock Differentiable distributionally robust optimization layers.
\newblock 2024.

\bibitem{DFL_survey_mandi2024decision}
Jayanta Mandi, James Kotary, et~al.
\newblock Decision-focused learning: Foundations, state of the art, benchmark
  and future opportunities.
\newblock {\em Journal of Artificial Intelligence Research}, 80:1623–1701,
  August 2024.

\bibitem{edge_resource_allocationmeskar2018fair}
Erfan Meskar and Ben Liang.
\newblock Fair multi-resource allocation with external resource for mobile edge
  computing.
\newblock In {\em INFOCOM WKSHPS}, pages 184--189, 2018.

\bibitem{michelmodeling}
Paul Michel, Tatsunori Hashimoto, and Graham Neubig.
\newblock Modeling the second player in distributionally robust optimization.
\newblock In {\em ICLR}, 2021.

\bibitem{DRO_mohajerin2018data}
Peyman Mohajerin~Esfahani and Daniel Kuhn.
\newblock Data-driven distributionally robust optimization using the
  wasserstein metric: Performance guarantees and tractable reformulations.
\newblock {\em Mathematical Programming}, 171(1):115--166, 2018.

\bibitem{patel2024splitwise}
Pratyush Patel, Esha Choukse, Chaojie Zhang, Aashaka Shah, Íñigo Goiri, Saeed
  Maleki, and Ricardo Bianchini.
\newblock Splitwise: Efficient generative llm inference using phase splitting.
\newblock In {\em 2024 ACM/IEEE 51st Annual ISCA}, pages 118--132, 2024.

\bibitem{DRL_environment_generation_ren2022distributionally}
Allen~Z. Ren and Anirudha Majumdar.
\newblock Distributionally robust policy learning via adversarial environment
  generation.
\newblock {\em IEEE Robotics and Automation Letters}, 7(2):1379--1386, 2022.

\bibitem{PPO_schulman2017proximal}
John Schulman, Filip Wolski, Prafulla Dhariwal, Alec Radford, and Oleg Klimov.
\newblock Proximal policy optimization algorithms.
\newblock 2017.

\bibitem{DDIM_song2020denoising}
Jiaming Song, Chenlin Meng, and Stefano Ermon.
\newblock Denoising diffusion implicit models.
\newblock 2022.

\bibitem{score-based_diffusion_SDE_song2021maximum}
Yang Song, Conor Durkan, Iain Murray, and Stefano Ermon.
\newblock Maximum likelihood training of score-based diffusion models.
\newblock 34:1415--1428, 2021.

\bibitem{score_matching_song2019generative}
Yang Song and Stefano Ermon.
\newblock Generative modeling by estimating gradients of the data distribution.
\newblock 32, 2019.

\bibitem{staib2017distributionally}
Matthew Staib and Stefanie Jegelka.
\newblock Distributionally robust deep learning as a generalization of
  adversarial training.
\newblock In {\em NeurIPS workshop on Machine Learning and Computer Security},
  volume~3, page~4, 2017.

\bibitem{stojkovic2024dynamollmdesigningllminference}
Jovan Stojkovic, Chaojie Zhang, Íñigo Goiri, Josep Torrellas, and Esha
  Choukse.
\newblock Dynamollm: Designing llm inference clusters for performance and
  energy efficiency, 2024.

\bibitem{policy_gradient_sutton1999policy}
Richard~S Sutton, David McAllester, Satinder Singh, and Yishay Mansour.
\newblock Policy gradient methods for reinforcement learning with function
  approximation.
\newblock 12, 1999.

\bibitem{AI_edge_computing_tuli2023ai}
Shreshth Tuli, Fatemeh Mirhakimi, Samodha Pallewatta, Giuliano Zawad, et~al.
\newblock Ai augmented edge and fog computing: Trends and challenges.
\newblock {\em Journal of Network and Computer Applications}, 216:103648, 2023.

\bibitem{Gen-DFL_wang2025gen}
Prince~Zizhuang Wang, Jinhao Liang, Shuyi Chen, Ferdinando Fioretto, and
  Shixiang Zhu.
\newblock Gen-dfl: Decision-focused generative learning for robust decision
  making.
\newblock 2025.

\bibitem{wang2024end}
Xinyu Wang, Yiyang Peng, and Wei Ma.
\newblock An end-to-end smart predict-then-optimize framework for vehicle
  relocation problems in large-scale vehicle crowd sensing.
\newblock {\em arXiv preprint arXiv:2411.18432}, 2024.

\bibitem{energy_consumption_LLM_heterogeneous_wilkins2024hybrid}
Grant Wilkins, Srinivasan Keshav, and Richard Mortier.
\newblock Hybrid heterogeneous clusters can lower the energy consumption of llm
  inference workloads.
\newblock In {\em Proceedings of the 15th ACM e-Energy}, pages 506--513, 2024.

\bibitem{wu2020proactive}
Jiajun Wu, Chengjian Sun, and Chenyang Yang.
\newblock Proactive optimization with machine learning: Femto-caching with
  future content popularity, 2020.

\bibitem{yang2024queueing}
Yuqing Yang, Lei Jiao, and Yuedong Xu.
\newblock A queueing theoretic perspective on low-latency llm inference with
  variable token length.
\newblock In {\em 2024 22nd WiOpt}, pages 273--280, 2024.

\bibitem{DL_workload_scheduling_ye2024deep}
Zhisheng Ye, Wei Gao, Qinghao Hu, Peng Sun, Xiaolin Wang, Yingwei Luo, Tianwei
  Zhang, and Yonggang Wen.
\newblock Deep learning workload scheduling in gpu datacenters: A survey.
\newblock {\em ACM Comput. Surv.}, 56(6), January 2024.

\bibitem{PTO_clustering}
Jinlei Zhang, Ergang Shan, Lixia Wu, Jiateng Yin, Lixing Yang, and Ziyou Gao.
\newblock An end-to-end predict-then-optimize clustering method for stochastic
  assignment problems.
\newblock {\em IEEE Transactions on Intelligent Transportation Systems},
  25(9):12605--12620, 2024.

\bibitem{data_center_demand_response_zhang2021hpc}
Yijia Zhang, Daniel~Curtis Wilson, Ioannis~Ch. Paschalidis, and Ayse~K. Coskun.
\newblock Hpc data center participation in demand response: An adaptive policy
  with qos assurance.
\newblock {\em IEEE Transactions on Sustainable Computing}, 7(1):157--171,
  2022.

\bibitem{risk_stochastic_opt_zhao2018data}
Chaoyue Zhao and Yongpei Guan.
\newblock Data-driven risk-averse stochastic optimization with wasserstein
  metric.
\newblock {\em Operations Research Letters}, 46(2):262--267, 2018.

\end{thebibliography}
\end{document}